\documentclass[10pt,twocolumn,letterpaper]{article}
\usepackage{cvpr}      
\usepackage[dvipsnames]{xcolor}

\definecolor{cvprblue}{rgb}{0.21,0.49,0.74}
\usepackage[pagebackref,breaklinks,colorlinks,citecolor=cvprblue]{hyperref}
\usepackage{makecell}
\usepackage{graphicx}
\usepackage{multirow}
\usepackage{amsmath}
\usepackage{algorithm}
\usepackage{algorithmicx}
\usepackage{algpseudocode}
\usepackage{colortbl}

\def\paperID{6968} 
\def\confName{CVPR}
\def\confYear{2025}

\title{TimeMarker: A Versatile Video-LLM for Long and Short Video Understanding with Superior Temporal Localization Ability}

\author{Shimin Chen\thanks{These authors contributed equally to this work.}, \quad Xiaohan Lan\footnotemark[1], \quad Yitian Yuan\footnotemark[1],\quad Zequn Jie,\quad Lin Ma\\
Meituan Inc. \\
{\tt \small chenshimin@zju.edu.cn \quad ruby\_lan\_123@outlook.com \quad  yuanyitian@foxmail.com }\\
{\tt \small zequn.nus@gmail.com \quad forest.linma@gmail.com }
}

\begin{document}
\maketitle
\begin{abstract}
Rapid development of large language models (LLMs) has significantly advanced multimodal large language models (LMMs), particularly in vision-language tasks. However, existing video-language models often overlook precise temporal localization and struggle with videos of varying lengths. We introduce TimeMarker, a versatile Video-LLM designed for high-quality dialogue based on video content, emphasizing temporal localization. TimeMarker integrates Temporal Separator Tokens to enhance temporal awareness, accurately marking specific moments within videos. It employs the AnyLength mechanism for dynamic frame sampling and adaptive token merging, enabling effective handling of both short and long videos. Additionally, TimeMarker utilizes diverse datasets, including further transformed temporal-related video QA datasets, to bolster its temporal understanding capabilities. Image and interleaved data are also employed to further enhance the model's semantic perception ability. Evaluations demonstrate that TimeMarker achieves state-of-the-art performance across multiple benchmarks, excelling in both short and long video categories. Our project page is at \url{https://github.com/TimeMarker-LLM/TimeMarker/}.
\end{abstract}

\section{Introduction}
\label{sec:intro}

\begin{figure}[ht]
\centering
\includegraphics[width=.95\columnwidth]{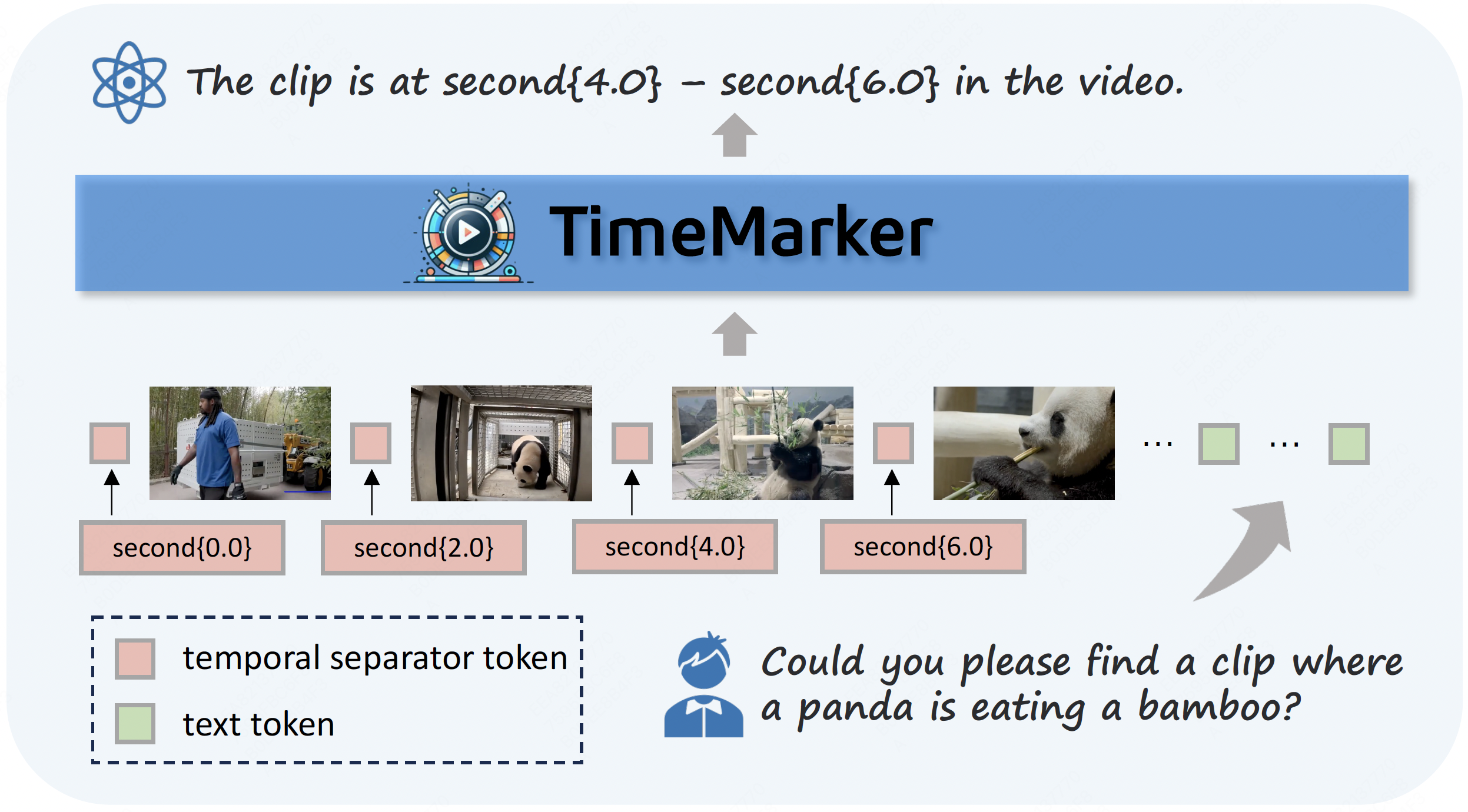}
\caption{The motivation of our proposed TimeMarker.  TimeMarker interleaves textual temporal separator tokens with video frame tokens, explicitly encoding the absolute temporal positions of video frames.}
\label{fig:motivation}
\end{figure}

The recent surge in the development of large language models (LLMs)~\cite{zeng2022glm,ray2023chatgpt,achiam2023gpt,touvron2023llama,dubey2024llama,cai2024internlm2} has significantly propelled the progress of multimodal large language models (LMMs)~\cite{li2023blip,zhu2023minigpt,bai2023qwenvl,chen2024internvl,liu2024visual,lin2023video,xu2024pllava,li2024llava}, especially in vision-language tasks. Videos, as rich sources of visual information, are crucial for understanding and interacting with real-world scenarios. However, current video-language models primarily focus on visual perception and reasoning, often overlooking the essential aspect of temporal localization and detection. This gap becomes evident when these models, despite being trained on extensive video captioning and question-answering (QA) datasets, fail to accurately place precise temporal references within video content~\cite{li2024videovista,wang2024lvbench,huang2024vtimellm,qian2024momentor}. The ability to search and locate specific moments within a video is critical for extracting relevant information from vast and redundant content and for providing accurate evidence-based answers to specific questions.

To enhance temporal perception in videos, some methodologies~\cite{zhang2023video,liu2024kangaroo,wang2024qwen2} incorporate temporal embeddings into video features, helping models understand the sequence of events. However, these approaches often focus on relative timing -- such as the order of events -- rather than absolute time points, such as the exact second an event occurs. This lack of precise temporal grounding (localization) can lead to less interpretable and verifiable responses, making subsequent temporal reasoning and inference more challenging.

In addition to the lack of absolute temporal perception, another challenge for video-LLMs is their ability to understand videos of different lengths. Videos expand information along the temporal dimension, causing the number of visual tokens to increase sharply with duration. Limited by the context length of LLMs or GPU capacity, some methods~\cite{maaz2023video,li2023videochat,lin2023video,luo2023valley} sample a small, fixed number of frames to extract features, focusing on short videos (e.g., within 3 minutes). Other models~\cite{lan2024vidcompress,he2024ma,li2025llama} attempt to handle longer videos (tens of minutes to hours) by sampling a greater and more flexible number of video frames, utilizing pooling or memory mechanisms to reduce the number of visual tokens per frame. However, this approach can result in the loss of detailed information. Consequently, developing a flexible method to effectively manage videos of varying lengths remains a significant challenge for video-LLMs.

To tackle these challenges, we present \textbf{TimeMarker}, a versatile Video-LLM crafted for high-quality dialogue based on content from videos of varying lengths, with a strong focus on temporal localization. Specifically, our TimeMarker has the following highlights:

\textbf{Temporal Separator Tokens Integration}:  TimeMarker enhances temporal awareness in videos by integrating Temporal Separator Tokens. This method interleaves textual temporal separator tokens (\textit{e.g.}, `second\{2.0\}' indicates that a particular video frame is at the 2.0 second mark of the video) with video frame tokens, explicitly encoding the absolute temporal positions of video frames. As shown in Figure~\ref{fig:motivation}, these tokens act as precise time markers, enabling the model to identify and reference specific moments within the video accurately.

\textbf{AnyLength Mechanism}: To accommodate videos of varying lengths, TimeMarker employs the AnyLength mechanism, which involves dynamic frame sampling and adaptive token resizing/merging. This mechanism adjusts the frames per second (FPS) and token compression ratio based on the video's length. For short videos, FPS is increased and token compression is reduced to capture more detail. For long videos, FPS is decreased and compression is increased to manage the extensive content efficiently. In addition, the above temporal separator tokens ensure that the LLM accurately interprets temporal information, even with dynamic changes in token count per frame.

\textbf{Advanced Data Utilization}: Beyond conventional video captioning and QA datasets, we also convert annotations from temporal action detection, segmentation, video summarization, and temporal sentence grounding into temporal-related video QA datasets. Temporal expressions are adapted to our tokenized format to enhance training on temporal tasks. Despite using only about 5M video-text pairs, our training videos span durations from under one minute to over 120 minutes. We also leverage about 85M images and 12M interleaved multi-image data. This diverse dataset boosts the model's semantic perception, cognitive abilities, and understanding of complex scenes.

\begin{figure}[!t]
\centering
\includegraphics[width=.95\columnwidth]{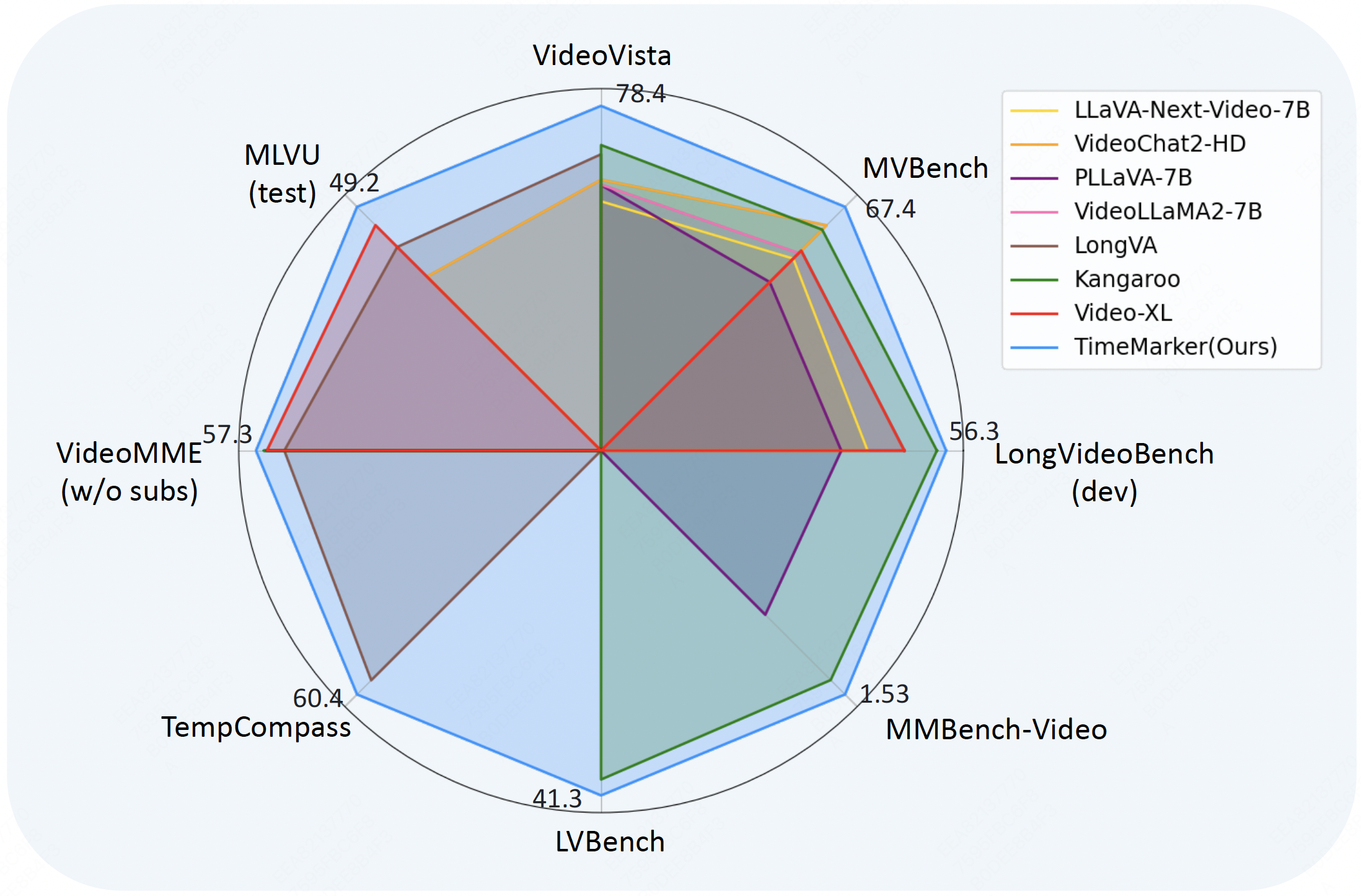}
\caption{TimeMarker attains leading performance across a range of comprehensive video understanding benchmarks. In the figure, we only list some 7B/8B models and highlight the specific scores of TimeMarker.}
\label{fig:radar}
\end{figure}

\textbf{Benchmark Excellence}: As shown in Figure~\ref{fig:radar}, TimeMarker achieves state-of-the-art performance across multiple public video benchmarks, excelling in both short and long video categories. It also surpasses traditional models in tasks such as temporal sentence grounding in videos, highlighting its superior temporal localization and understanding capabilities.

\begin{figure*}[!t]
\centering
\includegraphics[width=0.9\textwidth]{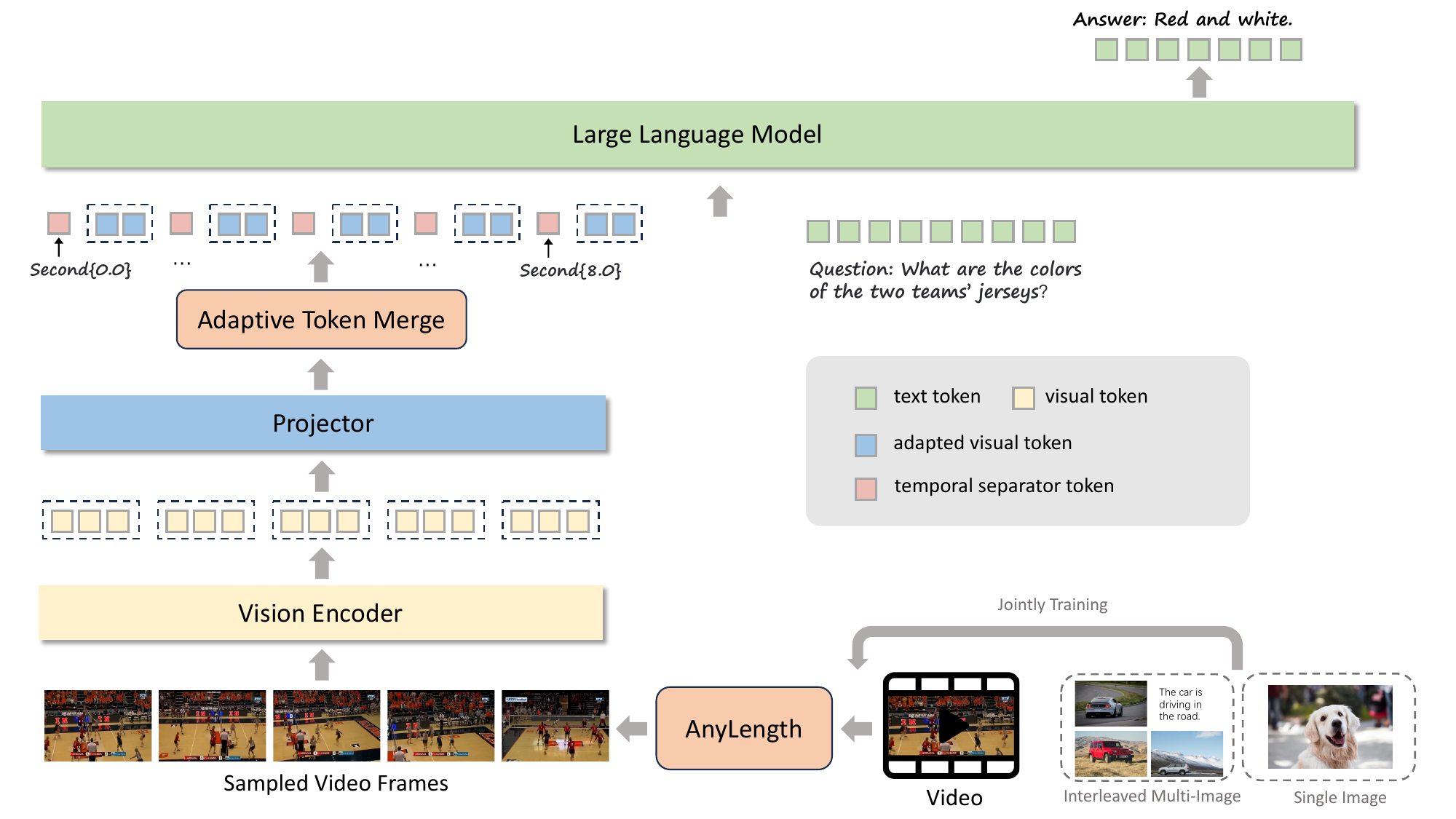}
\caption{The overview of our TimeMarker model. TimeMarker builds on the LLaVA architecture, using a Vision Encoder and cross-modality Projector to integrate visual and textual tokens for the LLM. We introduce an \textit{AnyLength} mechanism with an \textit{Adaptive Token Merge} module to handle varying video lengths, and \textit{Temporal Separator Tokens Integration} to encode temporal positions, enabling the LLM to perceive specific timestamps.}
\label{fig:framework}
\end{figure*}
\section{Related Work}
\label{sec:related_work}
\subsection{Large Image-language Models}
Benefiting from the success of large language models (LLMs)~\cite{zeng2022glm,ray2023chatgpt,achiam2023gpt,touvron2023llama,dubey2024llama,cai2024internlm2}, large image-language models have also made remarkable progress in vision-language understanding by incorporating image encoders into LLMs~\cite{wang2024qwen2,chen2024internvl,cai2024internlm2}. To mitigate the gap between visual and language modalities, BLIP~\cite{li2023blip} proposes to design a learnable query transformer (\ie, Q-Former) to act as a bridge between the frozen image encoder and LLM. LLaVA~\cite{liu2024visual} leverages a more lightweight linear layer to project visual tokens into the language space, advancing LLM with multimodal understanding abilities by fine-tuning both the projector and LLM. To better follow users' diverse instructions, MiniGPT-4~\cite{zhu2023minigpt} curates a high-quality, well-aligned dataset to increase conversational capabilities. Similarly, ShareGPT4V~\cite{chen2023sharegpt4v} constructs a large-scale dataset with detailed image descriptions to further enhance modality alignment.

\subsection{Large Video-language Models}
Video-language models~\cite{xu2024pllava,huang2024vtimellm,qian2024momentor,liu2024kangaroo} feed visual tokens of continuous video frames into an LLM, empowering it to comprehend videos. Compared to image-language models, video-based models take more effort on the temporal modeling. Early models~\cite{luo2023valley,maaz2023video,li2023videochat,lin2023video} typically use a limited and fixed number of frames as input. Video-LLaVA~\cite{lin2023video} that follows the structure of LLaVA uniformly samples 8 frames, using a pre-trained LanguageBind~\cite{zhu2023languagebind} to pre-align visual inputs. VideoChat2~\cite{li2023mvbench} adopts Q-Former to align a video encoder (\ie, UMT-L~\cite{li2023unmasked}) with LLM. These fixed-frame sampling methods fail to comprehend long videos (\eg, tens of minutes or hours). Some approaches~\cite{liu2024kangaroo,li2025llama,he2024ma} attempt to increase the number of sampled frames but have to decrease the number of visual tokens per frame due to the limited context length of LLMs. LLaMA-VID~\cite{li2025llama} introduces a dual-compression approach to transform each frame into two tokens, while MA-LMM~\cite{he2024ma} retains previous video content through a memory mechanism, feeding final visual tokens from the last iteration into the LLM. To reduce video tokens along both spatial and temporal dimensions, LongVU~\cite{shen2024longvu} designs an adaptive compression mechanism and Kangaroo~\cite{liu2024kangaroo} presents a 3D-conv-style patchify module. The compression manner of these methods, however, leads to inferior performance for short videos inevitably. Considering the concerns above, we devise an AnyLength mechanism to perform flexible sampling for videos of varying lengths.

\subsection{Temporal-awareness Video Understanding}
Temporal-related foundational vision tasks are crucial for evaluating a video model's temporal awareness and reasoning capabilities, representing an essential challenge in video content analysis.
For example, Video dense captioning~\cite{krishna2017dense,duan2018weakly} aims to describe continuous events occurring within a video, with each event corresponding to a pair of start and end timestamps. Temporal action detection~\cite{zhao2017temporal,vahdani2022deep}, on the other hand, focuses on locating all segments within a video where actions occur. Highlight detection~\cite{xiong2019less,lei2021detecting} involves identifying the most significant moments in a video, while temporal grounding~\cite{gao2017tall,yuan2019semantic,yuan2019find,zhang2020learning} refers to finding the specific segment in a video that corresponds to a given textual description.
Although current large video-language models~\cite{ren2023timechat,li2023videochat} incorporate various temporal-aware techniques to enhance temporal modeling, performance on most time-sensitive downstream video tasks remains suboptimal. To address this, we introduce Temporal Separator Tokens and curate QA pairs involving diverse temporal tasks for enhanced temporal awareness.

\section{TimeMarker}
\label{sec:method}

\begin{figure}[t] 
\centering
\resizebox{0.9\columnwidth}{!}{
\begin{minipage}{\columnwidth}
\begin{algorithm}[H] 
\begin{algorithmic}[1]
\Procedure{CalcKernelSize}{}
\State \textbf{Input: } $frame_N$ (the number of sampled frames), $kernel_b=2$ (the base kernel size)
\State \textbf{Output: } $kernel_h$ (kernel size of height for pooling), $kernel_w$ (kernel size of width for pooling)
\If{$frame_N > max\_frames/2$}
    \State $kernel_h \gets kernel_b*2$
    \State $kernel_w \gets kernel_b*2$
\ElsIf{$frame_N > max\_frames/4$}
    \State $kernel_h \gets kernel_b*2$
    \State $kernel_w \gets kernel_b$
\Else
    \State $kernel_h \gets kernel_b$
    \State $kernel_w \gets kernel_b$
\EndIf
\EndProcedure
\end{algorithmic}
\caption{Calculate Pooling Kernel Size}
\label{algorithm 2}
\end{algorithm}
\end{minipage}
}
\end{figure}

\subsection{Model Architecture and Pipeline}

Our TimeMarker is based on the fundamental architecture of LLaVA~\cite{liu2023llava}, which utilizes a Vision Encoder to process each video frame. The encoded visual tokens are then projected into the language space via a cross-modality Projector. These projected visual tokens are concatenated with input textual tokens and fed into the Large Language Model (LLM) to produce the final response. As depicted in Figure~\ref{fig:framework}, in addition to the basic components of LLaVA, we further propose \textbf{Temporal Separator Tokens Integration}, where these tokens are interleaved with video frame tokens as input to the LLM. This approach encodes the absolute temporal positions of video frames, allowing the LLM to explicitly recognize specific timestamps. Meanwhile, we also introduce an \textbf{AnyLength} mechanism with an \textbf{Adaptive Token Merge} module to enable our model to handle videos of varying lengths. Details of these new features in our TimeMarker will be provided in the following sections.

\begin{figure*}[!t]
\centering
\includegraphics[width=0.95\textwidth]{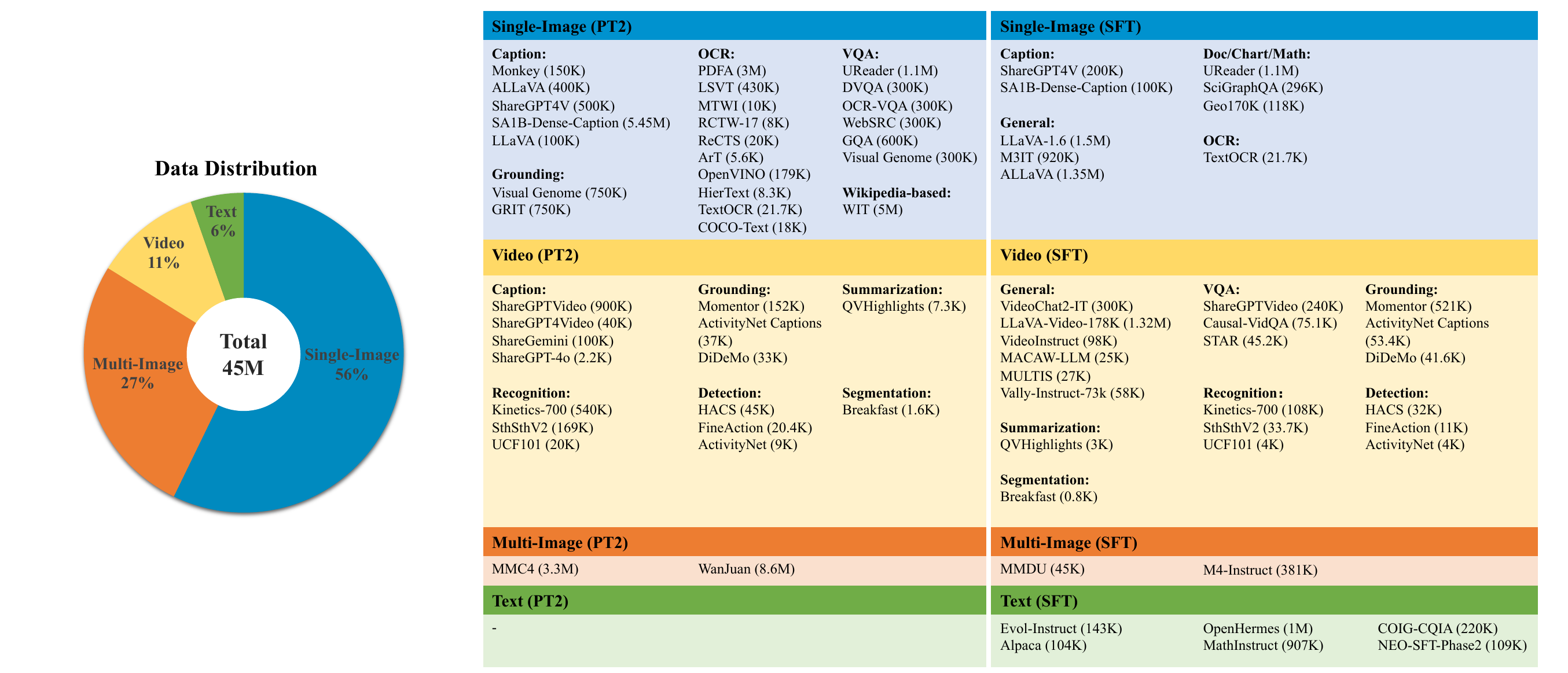}
\caption{The training data distribution and detailed list used in the PT2 and SFT stages of TimeMarker. In addition to video data, we utilize other multimodal data to assist in training TimeMarker: single images, interleaved multi-image, and pure text data.}
\label{fig:data}
\end{figure*}

\subsection{Temporal Separator Tokens Integration}

Expressing temporal information in video features is crucial for large video models, as precise temporal understanding impacts event comprehension. Previous large video models have either relied on the temporal encoding capabilities of LLMs or introduced learnable temporal embeddings attached to the visual features of video frames. However, these methods only perceive relative event order, not absolute time, such as the exact minute an event occurs, leading to suboptimal search and localization performance within videos.

To address this issue, our TimeMarker model not only utilizes the intrinsic sequential encoding capabilities of LLMs but also explicitly introduces \textbf{Temporal Separator Tokens} to help the video model perceive specific timestamps. As shown in Figure~\ref{fig:framework}, when sampling a video frame at second $i$, we prepend the text token ``$Second\{i\}$'' to its visual tokens before inputting them into the LLM. The sequence becomes ``$Second\{i\} || \mathbf{V}_i || Second\{j\} || \mathbf{V}_j $...'', where $Second\{i\}$ is the temporal separator token, $\mathbf{V}_i$ represents visual tokens of the video frame at second $i$, and $||$ denotes concatenation. Additionally, because we use the ``$Second\{i\}$'' format to express absolute time, any timestamp-related text in the training data is also converted into the same format as the temporal separator tokens. This helps the model achieve a unified understanding of timestamps.

Moreover, since temporal separator tokens are text tokens, it eliminates the need for realignment to the language space. This contrasts with methods using temporal embeddings, which necessitate aligning the learned embedding space to the language space, thereby complicating the LLM's understanding of temporal information. Additionally, our Temporal Separator Tokens are plug-and-play, making them highly adaptable to other model architectures.


\subsection{AnyLength Mechanism}

The main challenge in visual representation of images is managing varying input resolutions, which the Anyres strategy~\cite{li2024llava} aims to address. For videos, this challenge is greater due to their diverse durations, from seconds to hours. In video language models (VLMs), it's crucial to handle videos of any length while balancing performance and cost. The key factors affecting the number of visual tokens fed into the LLM are the number of sampled frames (video duration) and the number of tokens per frame (video resolution). We introduce an \textbf{AnyLength} mechanism to optimize these factors through dynamic frame sampling and adaptive token merging.

\noindent \textbf{Dynamic Video Frame Sampling in AnyLength.} The AnyLength mechanism samples video frames dynamically based on the input video duration ($dur$). Depending on the LLM's context length and GPU capacity, we first determine the maximum number of sampled frames in the video ($max\_frames$). The number of sampled frames per second  ($sample\_fps$) in video is then adjusted dynamically: For short videos of less than 8 seconds, we sample 2 frames per second to capture more visual details. For longer videos, we use a lower $sample\_fps = 1 / \lceil{dur / max\_frames}\rceil$, ensuring the total number of visual tokens does not exceed machine capacity, thus effectively preventing memory overflow during model training.

\noindent \textbf{Adaptive Token Merge in AnyLength.} After encoding the sampled video frames with the vision encoder and adapting them to the language space via the projector, we obtain a token feature map for each frame, denoted as $\mathbf{F} \in \mathbb{R}^{h \times w \times d}$, where $h$ and $w$ represent the number of tokens in the height and width dimensions, respectively. $d$ is the token dimension size. To further reduce the number of tokens per video frame and alleviate the model's overhead when processing long videos, we introduce an Adaptive Token Merge module. This module applies average pooling to $\mathbf{F}$ after the projector, with the kernel size adapted based on the actual number of video frames ($frame_{N}$) determined in the previous sampling step.

As outlined in Algorithm~\ref{algorithm 2}, we use a base kernel size of $kernel_b=2$. For longer videos where $frame_{N}$ exceeds certain thresholds, the kernel size is adjusted: if $frame_{N}$ is more than half of $max\_frames$, both height and width kernel sizes are doubled; if it is more than a quarter, only the height kernel size is doubled. Otherwise, both dimensions use the base size. In this case, we can further compress the video token numbers per frame based on the video duration.

\subsection{Training Strategy and Data}

To develop an efficient video-language model, we utilize a three-stage training pipeline as follows:

\noindent \textbf{PT1: Multimodal Alignment.}  
In this stage, we focus on aligning visual features with the language space by freezing the LLM and fine-tuning the vision encoder and projector. We use 60 million image-text pairs from datasets such as LAION-400M~\cite{schuhmann2021laion} (28.5\%), Zero~\cite{xie2023ccmb} (25\%), CapsFusion~\cite{yu2023capsfusion} (24\%), COYO-700M~\cite{kakaobrain2022coyo-700m} (20\%), CC12M~\cite{changpinyo2021conceptual} (1.5\%), and CC3M~\cite{sharma2018conceptual} (1\%). These datasets are recaptioned using InternVL2~\cite{chen2024internvl} to create high-quality image-text pairs.

\noindent \textbf{PT2: High-Quality Knowledge Learning.}  
This stage enhances the model's vision-language understanding through full-parameter training on diverse datasets that require detailed visual comprehension, video temporal understanding, and complex reasoning, moving beyond traditional methods~\cite{li2024llava} that mainly use image data. 

As shown in Figure~\ref{fig:data}, for single-image data, we incorporate caption, grounding, OCR, VQA, and wikipedia-based data. For video data, we include caption, recognition, and temporal-related data. We transform traditional action recognition data into multi-choice formats for video QA and use rule-based transformations to convert annotations from tasks like temporal action detection, segmentation, video summarization, and temporal sentence grounding into temporal-related video QA datasets. Additionally, we introduce multi-image datasets to enhance the model's understanding of image-text context, improving video comprehension.

\noindent \textbf{SFT: Instruction Tuning.}  
The final stage focuses on improving the model's ability to follow complex instructions, which is crucial for LLMs. We continue training all parameters of TimeMarker consistently with the PT2 stage. 

As shown in Figure~\ref{fig:data}, we employ recently proposed general instruction-following datasets for single-image, video, multi-image, and text. For image, besides caption, OCR and other general datasets, we also add complex QA datasets derived from documents, charts, and mathematical geometry. For video, we further incorporate complex reasoning VQA data. Similar to the PT2 stage, we convert annotations from various temporal-related tasks into temporal-related video QA datasets. However, in this stage, we reduce reliance on rule-based templates and instead enhance data complexity by additionally utilizing GPT-4o~\cite{gpt4o} to generate more intricate question-answer data.

Additionally, we also analyze the duration distribution of training videos in PT2 and SFT: 0.01\% are over 30 minutes, 4.9\% are 10-30 minutes, 6.6\% are 3-10 minutes, and 88.49\% are under 3 minutes. The longest video is 126 minutes. This demonstrates the diversity in our training video data durations.

\section{Experiments}
\label{sec:exp}

\begin{table*}[htbp]
  \centering
  \caption{Comparison of Video-LLMs on the short and general video benchmarks.}
  \resizebox{.8\textwidth}{!}{%
    \begin{tabular}{lllccccc}
    \toprule
    \makecell{\\Model Name} & \makecell{\\LLM} & \makecell{\\Frames} & \multicolumn{1}{p{5em}}{VideoMME (w/o subs)} & \multicolumn{1}{p{5em}}{VideoVista} & \multicolumn{1}{p{4.665em}}{MVBench} & \multicolumn{1}{p{5em}}{MMBench-Video} & \multicolumn{1}{p{6em}}{TempCompass} \\
    \midrule
    \multicolumn{8}{c}{Proprietary Models} \\
    \midrule
    Gemini-1.5-Pro~\cite{team2024gemini} & -     & 1/0.5 fps & \textbf{75.0}    & -  & -     & 1.30   & 67.1 \\
    GPT-4V~\cite{openai2023gpt4v} & -     &   10    & 59.9  & -     & 43.7  & 1.53  & - \\
    GPT-4o~\cite{hurst2024gpt} & -     & 384   & 71.9  & 78.3  & -     & \textbf{1.87}  & \textbf{73.7} \\
    Claude3-Opus~\cite{enis2024llm} & -     & 16    & 59.9    & -     & -     & -     & - \\
    \midrule
    \multicolumn{8}{c}{Open-Source Video MLLMs (\textgreater 8B)} \\
    \midrule
    VILA-1.5~\cite{lin2023vila} & 34B   & 14    & 62.3  & -     & 36.8     & 1.61  & - \\
    VideoLLaMA2-72B~\cite{cheng2024videollama} & 72B   & 32    & 62.4  & -     & 62.0    & -     & - \\
    Qwen2-VL-72B~\cite{wang2024qwen2} & 72B   & 2 fps   & 71.2  & -     & \textbf{73.6}  & -     & - \\
    LLaVA-Video~\cite{zhang2024video} & 72B   & 64    & 70.6  & -     & 62.8     & 1.71     & - \\
    InternVL-2~\cite{chen2024internvl} & 34B   & 16    & 61.2  & -     & -     & -     & - \\
    InternVL-1.5~\cite{chen2024internvl} & 20B   & 10    & 50.7  & -     & -     & -     & - \\
    VITA~\cite{fu2024vita}  & 8x7B  & 32    & 55.8    & -     & -     & -     & - \\
    \midrule
    \multicolumn{8}{c}{Open-Source Video MLLMs ($\leq$ 8B)} \\
    \midrule
    LLaVA-Next-Video~\cite{zhang2024llavanext-video} & 7B    & 32    & 33.7  & 56.7  & 53.1  & -     & - \\
    PLLaVA-7B~\cite{xu2024pllava} & 7B    &   16    & -     & 60.4  & 46.6  & 1.03  & - \\
    VideoChat2-HD~\cite{li2023mvbench} & 7B    & 16    & 45.6  & -     & 61.6  & 1.22  & - \\
    VideoLLaMA2-7B~\cite{cheng2024videollama} & 7B    & 16    & 47.9  & 60.5  & 54.6  & -     & - \\
    LongVA~\cite{zhang2024longva} & 7B    & 128   & 52.6  & 67.4  & -     & -     & 56.9 \\
    Video-XL~\cite{shu2024video} & 7B    & 128   & 55.5  & -     & 55.3  & -     & - \\
    Qwen2-VL-7B~\cite{wang2024qwen2} & 7B    &   2 fps    & 63.3  & 75.6     & 67.0    & -     & 67.8 \\
    LongVU~\cite{shen2024longvu} & 7B    & 1 fps  & -     & -     & 66.9  & -     & - \\
    Long-LLaVA~\cite{wang2024longllava} & 7B    & 64    & 52.9  & -     & -     & -     & - \\
    Kangaroo~\cite{liu2024kangaroo} & 8B    & 64    & 56.0    & 69.5  & 61.1  & 1.44  & - \\
    \rowcolor{gray!15}
    \textbf{TimeMarker~(Ours)} & 8B    & 128   & 57.3  & \textbf{78.4}  & 67.4  & 1.53  & 60.4 \\
    \bottomrule
    \end{tabular}%
    }
  \label{tab:general}%
\end{table*}%
\begin{table}[htbp]
  \centering
  \caption{Comparison of Video-LLMs on long video benchmarks.}
  \resizebox{\columnwidth}{!}{%
    \begin{tabular}{llcccc}
    \toprule
    \makecell{\\ Model Name} & \makecell{\\ LLM} & \multicolumn{1}{p{4.085em}}{VideoMME\newline{}-long(w/o subs) \newline{}\textit{30-60min}} & \multicolumn{1}{p{3.585em}}{LVBench\newline{}\textit{68min on aver.}} & \multicolumn{1}{p{4.085em}}{LongVideo\newline{}Bench (dev)\newline{}\textit{8s-60min}} & \multicolumn{1}{p{4.415em}}{MLVU (test) \newline{}\textit{3min-2hour}} \\
    \midrule
    \multicolumn{6}{c}{Open-Source Video MLLMs (\textgreater 8B)} \\
    \midrule
    VILA-1.5~\cite{lin2023vila} & 34B   & 53.8  & -     & -     & 44.2 \\
    VideoLLaMA2-72B~\cite{cheng2024videollama} & 72B   & 57.6  & -     & -     & 45.6 \\
    Qwen2-VL-72B~\cite{wang2024qwen2} & 72B   & \textbf{62.2} & \textbf{41.3}  & -     & - \\
    LLaVA-Video~\cite{zhang2024video} & 72B   & 61.5  & -     & -     & - \\
    InternVL-2~\cite{chen2024internvl} & 34B   & 52.6  & 39.6  & \textbf{59.3}     & 45.7 \\
    InternVL-1.5~\cite{chen2024internvl} & 20B   & 45.6  & -     & 51.2     & 37.3 \\
    VITA~\cite{fu2024vita}  & 8x7B  & 48.6  & -     & -     & - \\
    PLLaVA-34B~\cite{xu2024pllava} & 34B   & -     & 26.1  & 53.2  & - \\
    \midrule
    \multicolumn{6}{c}{Open-Source Video MLLMs ($\leq$ 8B)} \\
    \midrule
    LLaVA-Next-Video~\cite{zhang2024llavanext-video} & 7B    & -     & -     & 43.5  & - \\
    PLLaVA-7B~\cite{xu2024pllava} & 7B    & -     & -     & 40.2  & - \\
    VideoChat2-HD~\cite{li2023mvbench} & 7B    & -     & -     & -     & 35.1 \\
    VideoLLaMA2-7B~\cite{cheng2024videollama} & 7B    & -     & -     & -     & - \\
    LongVA~\cite{zhang2024longva} & 7B    & 46.2  & -     & -     & 41.1 \\
    Video-XL~\cite{shu2024video} & 7B    & 49.2  & -     & 49.5  & 45.5 \\
    LongVU~\cite{shen2024longvu} & 7B    & -     & -     & -     & - \\
    LongLLaVA~\cite{wang2024longllava} & 7B    & 45.4  & -     & -     & - \\
    Kangaroo~\cite{liu2024kangaroo} & 8B    & 46.6  & 39.4  & 54.2  & - \\
    \rowcolor{gray!15}
    \textbf{TimeMarker (Ours)} & 8B    & 46.4  & \textbf{41.3} & 56.3 & \textbf{49.2} \\
    \bottomrule
    \end{tabular}%
    }
  \label{tab:longbmk}%
\end{table}

\subsection{Implementation Details}

In our TimeMarker model, we use CLIP-ViT-L~\cite{radford2021learning} as the vision encoder with an input size of 336×336. The cross-modality projector, which aligns the vision encoder with the LLM, is a two-layer MLP~\cite{liu2024improved} with a GELU activation function. We select LLama3-8B~\cite{llama3} as the LLM. For PT1, we uniformly sample 16 frames from each video, while in later stages, we implement the AnyLength mechanism, with a maximum of 64 frames for PT2 and 128 for SFT. Our progressive training scheme uses a base learning rate of 1e-4 for PT1, 2e-5 for PT2, and 1e-5 for SFT, with a separate learning rate for ViT: 1e-5 for PT1 and 2e-6 for PT2 and SFT. We apply a layer-wise learning rate decay strategy~\cite{ishii2018layer} with a decay factor of 0.9 to effectively train ViT while maintaining its generalization performance.


\subsection{Benchmarks} 
We perform a comprehensive evaluation of TimeMarker's video understanding capabilities in three main aspects. For short and general video understanding, we utilize MVBench~\cite{li2023mvbench}, MMBench-Video~\cite{fang2024mmbench}, TempCompass~\cite{liu2024tempcompass}, VideoVista~\cite{li2024videovista}, and VideoMME~\cite{fu2024video} benchmarks. To assess long video understanding, we employ the long-video subset of VideoMME, along with LVBench~\cite{wang2024lvbench}, LongVideoBench~\cite{wu2024longvideobench}, and MLVU~\cite{zhou2024mlvu}. Additionally, we use temporal sentence grounding datasets Charades-STA~\cite{gao2017tall} and ActivityNet Captions~\cite{krishna2017dense} to examine the model’s temporal awareness and reasoning abilities.

\subsection{Results on Video Benchmarks}

\begin{table*}[ht]
\centering
\resizebox{1.0\textwidth}{!}{
\begin{tabular}{l|c|cccc|cccc}
\hline
\multicolumn{1}{l|}{\multirow{2}{*}{\textbf{Model Name}}} & \multicolumn{1}{c|}{\multirow{2}{*}{\textbf{Set up}}} & \multicolumn{4}{c|}{\textbf{Charades-STA}} & \multicolumn{4}{c}{\textbf{ActivityNet Captions}} \\
 & & R@1, IoU=0.3 & R@1, IoU=0.5 & R@1, IoU=0.7 & mIoU & R@1, IoU=0.3 & R@1, IoU=0.5 & R@1, IoU=0.7 & mIoU \\
\hline
2D-TAN~\cite{zhang2020learning} & FS & 57.3 & 45.8 & 27.9 & 41.0 & 60.4 & 43.4 & 25.0 & 42.5 \\
MMN~\cite{wang2022negative} & FS & 65.4 & 53.3 & 31.5 & 46.5 & 64.5 & 48.2 & 29.4 & 46.6 \\
UniVTG~\cite{lin2023univtg} & FS & 72.6 & \textbf{60.2} & \textbf{38.6} & \textbf{52.2} & - & - & - & - \\
\hline
Momentor~\cite{qian2024momentor} & VLM & 42.6 & 26.6 & 11.6 & 28.5 & 42.9 & 23.0 & 12.4 & 29.3 \\
ChatVTG~\cite{qu2024chatvtg} & VLM & 52.7 & 33.0 & 15.9 & 34.9 & 40.7 & 22.5 & 9.4 & 27.2 \\
VTimeLLM~\cite{huang2024vtimellm} & VLM & 55.3 & 34.3 & 14.7 & 34.6 & 44.8 & 29.5 & 14.2 & 31.4 \\
\rowcolor{gray!15}
\textbf{TimeMarker (Ours)} & VLM & \textbf{73.5} & 51.9 & 26.9 & 48.4 & \textbf{67.4} & \textbf{50.7} & \textbf{33.0} & \textbf{49.5} \\
\hline
\end{tabular}
}
\caption{\small Performance (\%) comparison of different models on Charades-STA and ActivityNet Captions benchmarks: FS denotes a model specialized for temporal sentence grounding in video with full supervision, while VLM refers to a Video-LLM. Notably, our TimeMarker operates in a zero-shot setting on Charades-STA, using no grounding data from Charades-STA during training.}
\label{tab:grounding_res}
\end{table*}
\begin{table}[!t]
\centering
\resizebox{\columnwidth}{!}{
\begin{tabular}{l|cc|cc}
\hline
\multicolumn{1}{l|}{\multirow{2}{*}{\textbf{Model Name}}} & \multicolumn{2}{c|}{\textbf{Charades-STA}} & \multicolumn{2}{c}{\textbf{ActivityNet Captions}} \\
 & R@1, IoU=0.7 & mIoU & R@1, IoU=0.7 & mIoU \\
\hline
TimeMarker & 26.9 & 48.4 & 33.0 & 49.5 \\
TimeMarker-wo-sep & 20.6 & 41.9 & 19.2 & 37.9\\
\hline
\end{tabular}
}
\caption{Examining the impact of temporal separator tokens on temporal sentence grounding in videos.}
\label{tab:grounding_ablation}
\end{table}

\noindent\textbf{Short and General Video Evaluation.} As shown in Table~\ref{tab:general}, we compare TimeMarker with other video-language models across various mainstream video benchmarks. For short video benchmarks like MVBench and MMBench-Video, TimeMarker surpasses all open-source models of similar size and competes well with larger proprietary models. This demonstrates that TimeMarker effectively retains key visual cues for short videos through its AnyLength mechanism and adaptive token merging strategy. Remarkably, on the comprehensive VideoVista benchmark, with videos ranging from a few seconds to over 10 minutes, TimeMarker achieves the highest score of 78.4, outperforming other open-source models and even rivaling GPT-4o. Additionally, on the VideoMME benchmark, which includes videos of varying lengths, and on TempCompass, which assesses video perception ability, TimeMarker shows strong performance comparable to models of similar size.


\noindent\textbf{Long Video Evaluation.}
Our TimeMarker model demonstrates significant strengths in handling long video content, performing competitively across key benchmarks, as shown in Table~\ref{tab:longbmk}. In LVBench, with videos averaging 68 minutes, TimeMarker achieves the highest score of 41.3, equaling the performance of larger models like Qwen2-VL-72B. On LongVideoBench, which features videos from 8 seconds to 60 minutes, TimeMarker scores 56.3, surpassing similar-size models such as Kangaroo (54.2) and even larger models like PLLaVA-34B (53.2). In the MLVU (test) benchmark, evaluating videos from 3 minutes to 2 hours, TimeMarker scores 49.2, outperforming models of the same or larger scale. Although its score on the VideoMME (long-video subcategory) benchmark is 46.4, which is lower than larger models like Qwen2-VL-72B (62.2) and VideoLLaMA2-72B (57.6), TimeMarker performs comparably to models of similar size.

The overall performance across multiple long-video benchmarks shows that, despite incorporating only a small amount of long video data into the training set, our devised AnyLength mechanism can effectively transfer the knowledge learned from short videos, highlighting TimeMarker as a versatile and effective solution for comprehensive long-video understanding.


\noindent\textbf{Summary.}
To summarize, the overall video benchmarking evaluation results demonstrate that its flexible AnyLength mechanism, adaptive token merging strategy, and injected temporal separator tokens are highly effective. These features allow TimeMarker to fully leverage the LLM context token capacity, showcasing exceptional comprehension capabilities across all short, medium, and long videos.

\begin{figure*}[!t]
\centering
\includegraphics[width=0.94\textwidth]{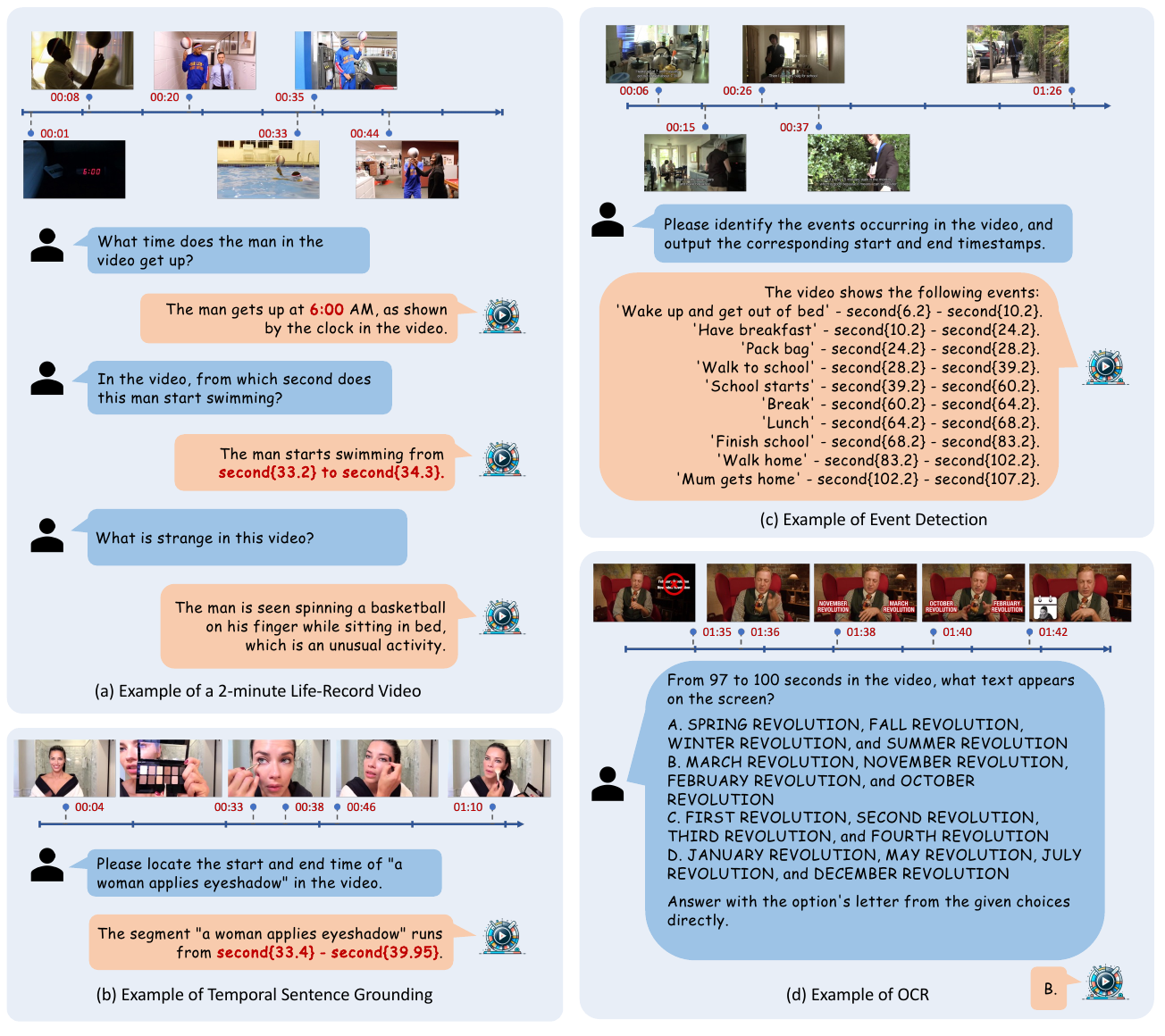}
\caption{Qualitative results across different scenarios. (a) multi-turn dialogue on a daily activity video, (b) temporal sentence grounding task in a beauty video, (c) event detection task in an interview video, and (d) OCR within a specific time interval.}
\label{fig:qua_res}
\end{figure*}

\subsection{Results on Temporal Sentence Grounding in Videos}

As shown in Table~\ref{tab:grounding_res}, TimeMarker excels in temporal sentence grounding tasks, significantly outperforming other VLM models on both benchmark datasets. In Charades-STA, which features shorter video sequences, TimeMarker achieves a top score of 73.5\% in R@1, IoU=0.3, surpassing the best fully supervised (FS) models, despite operating in a zero-shot setting that did not using any temporal sentence grounding data from Charades-STA in training. This highlights TimeMarker's superiority over traditional supervised methods. In the challenging ActivityNet Captions benchmark, with videos averaging 3 minutes, TimeMarker achieves the highest scores across all metrics. For example, it attains an R@1, IoU=0.7 score of 33.0\%, outperforming FS models like 2D-TAN (25.0\%) and MMN (29.4\%), as well as all other VLM models. It demonstrate TimeMarker's capability to handle longer and more complex videos for temporal semantic alignment.

\noindent \textbf{The Effects of Temporal Separator Tokens.} To assess the impact of temporal separator tokens on temporal sentence grounding, we test a degraded version of TimeMarker, called TimeMarker-wo-sep, which omits these tokens during both training and testing. As shown in Table~\ref{tab:grounding_ablation}, TimeMarker-wo-sep's performance significantly declines without the temporal separator tokens, underscoring their importance. These tokens provide explicit temporal cues, facilitating the LLM's ability to search and localize specific content within videos.

\subsection{Qualitative Results}

Figure~\ref{fig:qua_res} presents examples highlighting TimeMarker's superior temporal awareness and reasoning. In Figure~\ref{fig:qua_res}(a), a multi-turn dialogue from a 2-minute life-record video shows TimeMarker accurately identifying clock digits, locating relevant events, and reasoning about something strange. Figures~\ref{fig:qua_res}(b) and \ref{fig:qua_res}(c) demonstrate its performance in temporal sentence grounding and event detection. With Temporal Separator Tokens, TimeMarker shows a strong sense of absolute time, precisely localizing events and detecting boundaries in lengthy videos. Additionally, Figure~\ref{fig:qua_res}(d) showcases TimeMarker's ability to perform OCR tasks sequentially within a specified time interval.

\section{Conclusion}
\label{sec:conclusion}



In conclusion, TimeMarker provides an innovative solution to challenges in video-language models, particularly in temporal localization and handling videos of varying lengths. By integrating Temporal Separator Tokens and the AnyLength mechanism, it effectively encodes temporal positions and adapts to different video durations. Additionally, its advanced data utilization strategy enhances model training, resulting in superior performance across various video benchmarks. These innovations position TimeMarker as a leading model in temporal localization and understanding of videos, inspiring future advancements in video-LLMs.

\clearpage
{
    \small
    \bibliographystyle{ieeenat_fullname}
    \bibliography{main}
}

\end{document}